\documentclass{svproc}
\usepackage{graphicx}
\usepackage{marvosym}
\usepackage{amsmath}

\usepackage{url}

\usepackage{array}
\usepackage[noadjust]{cite}
\usepackage{latexsym,amssymb}
\usepackage{booktabs}
\usepackage[dvipsnames]{xcolor}

\usepackage[
    linkcolor=MidnightBlue,
    citecolor=MidnightBlue,
    filecolor=black,
    urlcolor=MidnightBlue,
]{hyperref}

\usepackage{adjustbox}

\newtheorem{construction}{Construction}
\newtheorem{heuristic}{Heuristic}

\newenvironment{proofsketch}{%
\proof}{\endproof}

\usepackage[ruled,vlined,linesnumbered]{algorithm2e}
\SetAlCapNameSty{textsc}
\SetFuncSty{texttt}
\SetFuncArgSty{ensuremath}
\SetArgSty{relax}
\SetNlSty{relax}{}{:}
\SetKw{KwBreak}{break}
\SetKw{KwFrom}{from}

\usepackage{booktabs}
\usepackage{longtable}
\newcommand*{\thead}[1]{%
\multicolumn{1}{c}{\bfseries\begin{tabular}{@{}c@{}}#1\end{tabular}}}

\def\orcidID#1{\unskip$^{[#1]}$}
\def\letter{$^{\textrm{(\Letter)}}$}

\begin{document}
\mainmatter              
\title{Efficient Parallel Algorithm for Decomposing Hard CircuitSAT Instances}
\titlerunning{Efficient Parallel \textnormal{CircuitSAT} Decomposition}  
%
\author{Victor Kondratiev\letter\orcidID{\href{https://orcid.org/0000-0003-0356-5149}{0000-0003-0356-5149}} \and\\
Irina  Gribanova\orcidID{\href{https://orcid.org/0000-0001-7155-4455}{0000-0001-7155-4455}} \and\\
Alexander Semenov\orcidID{\href{https://orcid.org/0000-0001-6172-4801}{0000-0001-6172-4801}}}

\authorrunning{V. Kondratiev \textit{et al.}} 
%
\tocauthor{Victor Kondratiev, Irina Gribanova, and Alexander Semenov}
\institute{ISDCT SB RAS, Irkutsk, Russia \\
	\email{\href{mailto:vikseko@gmail.com}{vikseko@gmail.com}}
 \\
}
	
\maketitle              

\begin{abstract}
We propose a novel parallel algorithm for decomposing hard \textnormal{CircuitSAT} instances.
The technique employs specialized constraints to partition an original \textnormal{SAT} instance into a family of weakened formulas.
Our approach is implemented as a parameterized parallel algorithm, where adjusting the parameters allows efficient identification of high-quality decompositions, guided by hardness estimations computed in parallel.
We demonstrate the algorithm's practical efficacy on challenging \textnormal{CircuitSAT} instances, including those encoding \textup{Logical Equivalence Checking} of Boolean circuits and preimage attacks on cryptographic hash functions.


\end{abstract}

\section{Introduction}

Digital circuits form the foundation of the modern world, from microprocessors to specialized electronics (e.g., \textup{FPGA}, \textup{ASIC}).
As a mathematical model of combinational digital circuits, Boolean circuits are particularly convenient because they capture most logical and structural properties.
They enable the study of mathematical and combinatorial characteristics of digital circuits and facilitate the development of testing and verification algorithms.
Due to the high complexity of modern circuits, they are typically designed using specialized tools in the field of electronic design automation (\textup{EDA}).

%
%

The \textup{Logical Equivalence Checking} (\textup{LEC}) problem~\cite{molitor2004} is critically important in \textup{EDA}.
Modern \textup{EDA} tools typically employ complete \textnormal{SAT} solvers~\cite{HandBook09} to address \textup{LEC}.
The applications of \textnormal{SAT} solvers have expanded significantly over the past 25 years, particularly since the introduction of the conflict-driven clause learning (CDCL) algorithm~\cite{GRASP1996, GRASP1999, DBLP:series/faia/0001LM21}.
Today, \textup{CDCL}-based \textnormal{SAT} solvers are widely used in verification and software analysis~\cite{biere1999a, DBLP:series/faia/Kroening21}, cryptography~\cite{bard2009, DBLP:conf/pact/SemenovZBP11, AAAI2018}, planning~\cite{Rintanen2021}, combinatorics~\cite{Zhang2021}, and as sub-engines in solvers from \textup{satisfiability modulo theories} (SMT)~\cite{deMoura2008}.

However, reducing \textup{LEC} to \textnormal{SAT} often produces instances that are extremely challenging for state-of-the-art \textnormal{SAT} solvers, motivating the development of parallel-solving methods.
Several strategies exist for parallel \textnormal{SAT} solving. In this paper, we employ the partitioning strategy~\cite{Hyvarinen2011}, specifically a variant tailored to \textnormal{SAT} instances derived from Boolean circuits.

The main contributions of our work are as follows:
\begin{enumerate}
    \item We present a novel parallel algorithm for solving a broad class of \textnormal{CircuitSAT} problems, leveraging a partitioning construction introduced in~\cite{IEEE2025}.

    \item We implement the algorithm as a parallel \textup{MPI} application.

    \item We apply our tool to solve hard \textup{LEC} instances and algebraic cryptanalysis problems that are intractable for even the best sequential \textnormal{SAT} solvers, demonstrating its practical utility.
\end{enumerate}
The paper is organized as follows. 
Section 2 introduces the basic concepts, terms, and notation used throughout the paper.  
Section 3 details the method for constructing SAT partitionings based on pseudo-Boolean intervals.  
Section 4 presents a novel parallel algorithm designed for solving a broad class of Circuit-SAT problems.  
Section 5 reports the results of computational experiments in which we evaluated the proposed methods in application to LEC problems for sorting algorithms and inversion problems for the MD4 function.
In the concluding section we  summarize the obtained results and outline possible directions for further work.

\section{Preliminaries}
Boolean circuits are directed graphs that specify discrete functions of the form \( f \colon \{0,1\}^n \rightarrow \{0,1\}^m \).
Let \( G_f \) be such a graph with a set of vertices \( \mathcal{V} \) and a set of arcs \( \mathcal{A} \).
The set \( \mathcal{V} \) contains two subsets: (1) a set of \( n \) parentless vertices called the \textit{circuit inputs}, and (2) a set of \( m \) vertices called the \textit{circuit outputs}.
Each vertex that does not belong to the set of inputs is associated with an element of a complete basis~\cite{wegener1987}, representing an elementary Boolean function.
These basis elements are called \textit{gates}.
If the basis \( \{\lnot, \land\} \) is used, the resulting graph is called an \textit{And-Inverter Graph} (AIG)~\cite{Kuehlmann2002}.
We refer to \( G_f \), labeled in this way, as a \textit{Boolean circuit} and denote it by \( S_f \).

Given an input vector \( \alpha \in \{0,1\}^n \), the \textit{interpretation} of \( S_f \) on \( \alpha \) is the sequence of computations of the elementary Boolean functions at each gate.
The output of this interpretation is the vector formed by the values at the output gates.
Thus, \( S_f \) defines a total function \( f \colon \{0,1\}^n \rightarrow \{0,1\}^m \).

One of the central problems in EDA is \textit{Logical Equivalence Checking} (LEC), which verifies whether two Boolean circuits \( S_f \) and \( S_g \) specify the same function, i.e., whether \( f \cong g \) (point-wise equivalence).

The standard approach to LEC involves \textnormal{SAT} solvers.
Recall that the \textit{Boolean satisfiability problem} (SAT) asks whether a given Boolean formula is satisfiable.
An \textit{assignment} is any set of values for the variables in the formula.
A formula is \textit{satisfiable} if there exists an assignment for which it evaluates to \( 1 \) (True). If a formula evaluates to $0$ (False) for any assignment, it is said to be \textit{unsatisfiable}.

Tseitin transformations~\cite{tseitin1970} provide a polynomial-time reduction from the satisfiability problem of an arbitrary Boolean formula to the satisfiability problem for a formula in conjunctive normal form (CNF), i.e., a conjunction of elementary disjunctions called \textit{clauses}.
As a result of this reduction, SAT is most commonly considered as the problem of determining the satisfiability of an arbitrary CNF.

By applying Tseitin transformations to a Boolean circuit $S_f$, we can construct a CNF $C_f$ in polynomial time relative to the number of gates in $S_f$.
Following~\cite{semenov2020}, we refer to $C_f$ as a \textit{template CNF}.
Algorithms for constructing $C_f$ from $S_f$ are well-known and described in, e.g.,~\cite{semenov2020, IEEE2025}.
Various properties of Boolean circuits can be analyzed by applying SAT solvers to formulas derived from template CNFs.

Let $f \colon \{0,1\}^n \rightarrow \{0,1\}$ be an arbitrary total Boolean function, and let $S_f$ be a Boolean circuit implementing $f$.
The \textnormal{CircuitSAT} problem asks whether there exists an input to $S_f$ for which the circuit outputs $1$ (True).
Using Tseitin transformations, this problem can be reduced (in linear time relative to the number of gates in $S_f$) to \textnormal{SAT} for a CNF obtained by augmenting the template CNF $C_f$ with a clause consisting of one variable.

Consider two Boolean circuits $S_f$ and $S_g$, $f,g\colon \{0,1\}^n \rightarrow \{0,1\}^m$.
Given $S_f$ and $S_g$, we construct a \textit{miter circuit} (see~\cite{molitor2004, IEEE2025}) and then derive a CNF $C_{f \oplus g}$.
This CNF is unsatisfiable if and only if $S_f$ and $S_g$ are equivalent.
The transformation from $S_f$ and $S_g$ to $C_{f \oplus g}$ takes linear time relative to the total number of gates in $S_f$ and $S_g$.
Thus, the LEC problem for $S_f$ and $S_g$ can be efficiently reduced to SAT for $C_{f \oplus g}$.

Recall that SAT is a well-known NP-complete problem~\cite{cook1971,Karp1972}, meaning it cannot be solved in polynomial time (in the general case) if it is assumed that $P \neq NP$.
However, many special cases involving large formulas (with tens or hundreds of thousands of variables and clauses) can be solved efficiently using algorithms based on simple, natural techniques.
In particular, SAT solvers implementing the CDCL algorithm~\cite{DBLP:series/faia/0001LM21} perform well on various industrial problems, including LEC instances.

Nevertheless, many combinatorial problems produce extremely hard SAT instances when reduced to SAT.
For LEC, this occurs with circuits implementing arithmetic functions such as integer multiplication or sorting algorithms for sets of natural numbers.
Even the best modern SAT solvers cannot solve such relatively small-scale instances in a reasonable time, and their runtime behavior in these cases remains unpredictable. This motivates the problem of estimating formula hardness relative to a specific algorithm (e.g., a SAT solver).
One approach is to decompose the original formula into a family of simpler subproblems with a relatively small solver runtime.
The total solving time across all subproblems in the decomposition then provides an upper bound on the hardness of the original instance.
This approach has been developed in~\cite{CP2021,SprOpt2021,IJAI2023,PACT2022}, among others.
Below, we outline its core idea.

Let $C$ be an arbitrary CNF over a set $X$ of variables, and let $A$ be a complete SAT-solving algorithm (e.g., a CDCL-based solver).
We refer to a Boolean variable $x \in X$ or its negation $\lnot x$ as a \textit{literal}. Following~\cite{Szeider}, we use the notation
$$
 x^{\sigma} = \begin{cases}
        x, & \sigma = 1,\\
        \lnot x, & \sigma = 0.
    \end{cases}
$$
Assume $|X| = k$; the set of all assignments to $X$ then forms a Boolean hypercube $\{0,1\}^k$.
For an arbitrary subset $B \subseteq X$, the assignments to $B$ similarly form a hypercube $\{0,1\}^{|B|}$.

For an arbitrary subset $B \subseteq X$ and assignment $\beta \in \{0,1\}^{|B|}$, the substitution of $\beta$ into formula $C$ is performed in the standard way (see, e.g., \cite{chang1973}).
The resulting formula is denoted by $C[\beta/B]$.
For any $B \subseteq X$, we associate the set of $2^{|B|}$ possible formulas of the form $C[\beta/B]$, where $\beta \in \{0,1\}^{|B|}$.

Let $t_A(C)$ denote the runtime of SAT solver $A$ on formula $C$.
Following \cite{CP2021}, we define the \textit{decomposition hardness} of $C$ with respect to algorithm $A$ and decomposition set $B$ as
\begin{equation}
\label{eq1}
\mu_{A,B}(C) = \sum_{\beta \in \{0,1\}^{|B|}} t_A(C[\beta/B]).
\end{equation}
Here, the key quantity of interest is
\begin{equation}
\label{eq2}
\mu_A(C) = \min_{B \subseteq X} \mu_{A,B}(C).
\end{equation}

Both \eqref{eq1} and \eqref{eq2} serve as upper bounds on the hardness of $C$, since there exists an algorithm that solves SAT for $C$ by applying a complete SAT solver to all formulas $C[\beta/B]$.
The computation of \eqref{eq2} can be approached using black-box optimization algorithms.
We emphasize that the set $\{C[\beta/B] \mid \beta \in \{0,1\}^{|B|}\}$ can be processed in parallel.

This approach has proven effective in several cases, enabling, for example, the construction of non-trivial attacks on certain cryptographic functions (see \cite{SprOpt2021}).
However, for hard LEC instances, more specialized constructions often yield better results. We present one such construction in the next section.

\section{Construction of SAT Partitioning for CircuitSAT}
\label{sec3}
The concept of SAT partitioning was introduced in~\cite{Hyvrinen2006} (see also~\cite{Hyvarinen2011}).

\begin{definition}[\cite{Hyvarinen2011}]
Let $C$ be a CNF formula over a set of Boolean variables $X$, and let $\Pi = \{G_1, \ldots, G_s\}$ be a set of Boolean formulas over $X$.
The set $\Pi$ is called a \emph{SAT partitioning} of $C$ if the following conditions hold:
\begin{enumerate}
    \item formulas $C$ and $C \land (G_1 \lor \cdots \lor G_s)$ are equisatisfiable;
    \item for all distinct $i, j \in \{1, \ldots, s\}$, the formula $C \land G_i \land G_j$ is unsatisfiable.
\end{enumerate}
\end{definition}

The construction that divides $C$ into formulas of the form $C[\beta/B]$ for some $B \subseteq X$ trivially yields a SAT partitioning of $C$.
Below, we employ an alternative partitioning construction from~\cite{IEEE2025}, specifically designed for CircuitSAT problems.

Consider a CircuitSAT problem, which could represent either an LEC problem or an inversion problem for a function $f \colon \{0,1\}^n \rightarrow \{0,1\}^m$ implemented by a circuit $S_f$.
Given $\gamma \in \mathrm{Range}(f) \subseteq \{0,1\}^m$, the inversion problem requires finding $\alpha \in \{0,1\}^n$ such that $f(\alpha) = \gamma$.

Let $C$ be the CNF encoding this CircuitSAT problem, which we refer to as the \textit{associated CNF}.
Let $X$ denote the set of variables in $C$, and let $X^{\mathrm{in}} = \{x_1, \ldots, x_n\}$ be the subset of $X$ corresponding to inputs of circuit $S_f$.
We interpret each assignment $\alpha \in \{0,1\}^{|X^{\mathrm{in}}|}$ as the coefficients of a binary number in $N_0^n = \{0, \ldots, 2^n - 1\}$, establishing a bijection $\phi \colon \{0,1\}^n \rightarrow N_0^n$.

For $a, b \in N_0^n$, the set $\{p \in N_0^n \mid a \leq p < b\}$ is called an \textit{interval}, denoted by $[a, b)$, with length $b - a$.
The corresponding set of Boolean vectors satisfies the integer inequality
\begin{equation}
\label{eq3}
a \leq x_1 + 2x_2 + \cdots + 2^{n-1}x_n < b,
\end{equation}
where $x_i \in \{0,1\}$, $i \in \{1, \ldots , n\}$.

\begin{definition}
A set $\mathcal{R}^n$ of intervals is called a \emph{complete system of intervals} if it satisfies the following conditions:
\begin{enumerate}
    \item all intervals are pairwise disjoint;

    \item every number $\alpha \in N_0^n$ belongs to some interval in $\mathcal{R}^n$.
\end{enumerate}
\end{definition}

Any interval $I = [a, b) \in \mathcal{R}^n$ with $b - a \geq 2$ can itself be decomposed into a complete system of intervals, which we denote by $\mathcal{R}(I)$.

The following construction was introduced in~\cite{IEEE2025}.

\begin{construction}[\cite{IEEE2025}]
Let $\mathcal{R}^n$ be a complete system of intervals. For each $I = [a, b) \in \mathcal{R}^n$, associate the inequality \eqref{eq3} and let $C_I$ be the CNF produced by this inequality using standard translation methods~\cite{EenSorr2006}. We say that $C_I$ \emph{encodes} the interval $I$. Define $\Pi = \{C_I\}_{I \in \mathcal{R}^n}$.
\end{construction}

\begin{theorem}[\cite{IEEE2025}]
\label{ieeethm}
For any CNF $C$ encoding a CircuitSAT problem and any complete interval system $\mathcal{R}^n$, the set $\Pi = \{C_I\}_{I \in \mathcal{R}^n}$ forms a SAT partitioning of $C$.
\end{theorem}

This \textit{interval partitioning} method has proven effective for solving
challenging LEC problems~\cite{IEEE2025}. In the next section, we present a novel algorithm for constructing SAT partitionings based on the described construction.

\section{Adaptive Parallel Algorithm for Constructing Partitionings in CircuitSAT}
\label{sec4}
We present a new algorithm for building SAT partitionings for CircuitSAT instances using the interval construction described above. The section is divided into two parts: the first describes the basic algorithm and provides a proof of its completeness, while the second introduces additional heuristics that improve the practical performance of the basic algorithm.

\subsection{Basic Algorithm}
\label{subsec:basicalg}
Consider a CNF $C$ over variables $X$, constructed for a Boolean circuit $S_f$ implementing $f \colon \{0,1\}^n \rightarrow \{0,1\}^m$. Accordingly, the set $X$ contains a subset $X^{\mathrm{in}}$, $|X^{\mathrm{in}}| = n$, consisting of variables assigned to the inputs of $S_f$.

The algorithm uses a CDCL SAT solver $A^t$ whose running time is limited by a constant $t$. Such a solver is a polynomial sub-solver in the sense of~\cite{Williams2003}. On input CNF $C$, $A^t$ returns one of three possible outputs:
\begin{itemize}
    \item SAT ($C$ is satisfiable);
    \item UNSAT ($C$ is unsatisfiable);
    \item INDET (satisfiability cannot be determined within time $t$).
\end{itemize}
The time limit $t$ can be expressed either as physical time (in seconds) or as the number of elementary operations performed by the solver. For CDCL SAT solvers, it is convenient to limit the runtime by the number of conflicts~\cite{DBLP:series/faia/0001LM21}.

The algorithm involves several parameters. The first parameter, $q$, represents the size of the initial interval partitioning.
At the initial step, a complete interval system $\mathcal{R}^n$ consisting of $q$ intervals is constructed. We denote this system by $\mathcal{R}_0^n = \{I_1, \ldots, I_q\}$. Each interval $I_j$, $j \in \{1, \ldots, q\}$, is associated with its CNF encoding $C_0^j$. The value of $q$ is selected based on the computing environment's capabilities; for example, $q$ may equal the number of available computing cores to which the formulas $C_0^j$, $j \in \{1, \ldots, q\}$, are distributed.

For each $j \in \{1, \ldots, q\}$, the SAT solver $A^t$ is applied to the formula $C \land C_0^j$. If $A^t(C \land C_0^j)$ returns INDET, then $C_0^j$ encodes an interval $I_j \in \mathcal{R}_0^n$ that can be divided into smaller intervals. For simplicity, we assume each subsequent partition splits into at most $d$ intervals, where $d$ is another pre-specified algorithm parameter.

The process terminates in one of two cases:
\begin{enumerate}
    \item When $A^t$ finds a satisfying assignment for some CNF $C(I)$ encoding interval~$I$.
    \item When the unsatisfiability of $C \wedge C(I)$ for every interval $I$ generated during the algorithm's execution was proved.
\end{enumerate}

The algorithm's operation is represented as a tree $T(C)$ with $q$ branches from the root. The root is associated with the pair $(C, \mathcal{R}_0^n)$, and its branches correspond to the formulas $C_0^j$ for $j \in \{1, \ldots, q\}$. If $A^t$ finds a satisfying assignment for $C_0^j \land C$, then the algorithm terminates with a solution for $C$. If $A^t$ proves $C_0^j \land C$ is unsatisfiable, then the corresponding child node in the tree $T(C)$ is marked with $\bot$ and becomes a leaf.

When $A^t$ returns INDET for $C_0^j \land C$, the child node is associated with $(C, \mathcal{R}(I_j))$, where $\mathcal{R}(I_j)$ divides $I_j$ into shorter intervals. This node then becomes the new root for further recursive processing.

Let $v$ be an arbitrary vertex in the tree $T(C)$ at depth $l$, where $l$ equals the number of edges from the root to $v$ (thus $l=0$ for the root). We will use the term ``vertex at level $l$'' to refer to $v$.

\begin{algorithm}[h!]
\caption{DFS-based CircuitSAT decomposition}%
\label{algo:dfs-decomp}

\DontPrintSemicolon
\KwIn{
    CNF formula $C$, number of initial intervals $q$, splitting factor $d$, timeout $t$
}
\KwOut{
    SAT (satisfying assignment) or UNSAT
}
\BlankLine
Let $\mathcal{R}_0^q = \{I_1, \dots, I_q\}$ be the initial set of intervals\;
Initialize an empty LIFO queue $Q$\;
\ForEach{$j \in \{1, \dots, q\}$}{
    Construct CNF $C_j$ encoding interval $I_j$\;
    Push $(C \land C_j, I_j)$ onto $Q$\;
}
\BlankLine
\While{$Q$ is not empty}{
    Pop $(C_{\mathrm{current}}, I_{\mathrm{current}})$ from $Q$\;
    Run SAT solver $A^t$ on $C_{\mathrm{current}}$ with timeout $t$\;
    \eIf{$A^t$ returns SAT}{
        \Return{SAT} \tcp{Return the satisfying assignment}
    }{
        \If{$A^t$ returns UNSAT}{
            \textbf{Continue} \tcp{Discard the current interval}
        }
        \Else{
            Split $I_{\mathrm{current}}$ into $d$ subintervals $\{I_1', \dots, I_d'\}$\;
            \ForEach{$i \in \{1, \dots, d\}$}{
                Construct CNF $C_i'$ encoding interval $I_i'$\;
                Push $(C \land C_i', I_i')$ onto $Q$\;
            }
        }
    }
}
\BlankLine
\Return{UNSAT}\;
\end{algorithm}

Our goal is to prove that under certain general conditions on $A^t$, the described procedure will find a satisfying assignment for satisfiable CNF $C$ and construct a finite tree with $\bot$-labeled leaves for unsatisfiable $C$. We preface this result with the following lemma.

\begin{lemma}
\label{lemma1}
Consider a Boolean circuit $S_f$ specifying a total function $f \colon \{0,1\}^n \rightarrow \{0,1\}^m$, and let $C_f$ be its template CNF in the sense of~\cite{semenov2020}.
Let $X$ denote the variables in $C_f$, with $X^{\mathrm{in}} = \{x_1, \ldots, x_n\}$ being the variables from $X$ associated with the inputs of $S_f$.
For any input $\alpha = (\alpha_1, \ldots, \alpha_n) \in \{0,1\}^n$, consider the CNF
\begin{equation}
\label{eq4}
x_1^{\alpha_1} \land \cdots \land x_n^{\alpha_n} \land C_f.
\end{equation}
Applying only the Unit Propagation rule~\cite{DBLP:series/faia/0001LM21} to \eqref{eq4} derives values for all variables in $X$ (as literals) without conflicts.
For variables $y_1, \ldots, y_m$ associated with outputs of $S_f$, this yields $y_1 = \gamma_1, \ldots, y_m = \gamma_m$, where $f(\alpha) = \gamma$ and $\gamma = (\gamma_1,\ldots,\gamma_m)$.
\end{lemma}

This lemma is well-known and appears in several contemporaneous works~\cite{bessiere2009, jarvisalo2009, Semenov2009}.
Its validity follows directly from the properties of Tseitin transformations.

We now prove the completeness of the tree construction algorithm for $T(C)$ described above.

\begin{theorem}
\label{thm1}
Let $A^t$ be a CDCL-based sub-solver that performs at most $t$ conflicts for any $t \geq 1$. Then the algorithm for constructing $T(C)$ terminates after finitely many calls to $A^t$.
\end{theorem}

\begin{proofsketch}
Consider any internal vertex $v$ of $T(C)$ associated with a pair $(C, \mathcal{R}(I))$ for some interval $I \in \mathcal{R}^n$. Let $\mathcal{R}(I) = \{I_1', \ldots, I_p'\}$ be a partition of $I$. Assume that each $I' \in \mathcal{R}(I)$ has length 1 and its unique number has binary representation $\alpha = (\alpha_1, \ldots, \alpha_n)$. Obviously, the CNF encoding $I'$ becomes the conjunction of literals
\[
x_1^{\alpha_1} \land \cdots \land x_n^{\alpha_n}.
\]

In this case, $A^t$ receives $x_1^{\alpha_1} \land \cdots \land x_n^{\alpha_n} \land C$ as input and outputs all variable values in $C$ according to Lemma~\ref{lemma1}. If $C$ contains additional constraints beyond the template $C_f$ (e.g., miter circuit constraints for LEC~\cite{molitor2004, IEEE2025}), then $A^t$ will either prove that $C$ is unsatisfiable or
find a satisfying assignment (terminating the algorithm).

Since partitioning always decreases interval lengths, any interval $I = [a, b)$ splits into intervals of length less than $b - a$. This means that after a finite number of partitioning steps, any interval will eventually be partitioned into intervals of length 1.

It follows from the above that, for each such interval $I'$, the algorithm $A^t$ solves the satisfiability problem for CNF $C(I') \land C$ with at most one conflict. In the worst case, the tree $T(C)$ has $2^n$ leaves, each requiring one $A^t$ call on a CNF of the form $x_1^{\alpha_1} \land \cdots \land x_n^{\alpha_n} \land C$. Thus, Theorem~\ref{thm1} holds.
$\blacksquare$
\end{proofsketch}

The algorithm's implementation traverses $T(C)$ using depth-first search (DFS). Each branch ending in a $\bot$-labeled leaf is pruned. During construction, if $A^t$ proves satisfiability for any CNF $C(I) \land C$, the algorithm terminates; if all branches are pruned, $C$ is unsatisfiable. The corresponding pseudocode is given in Algorithm~\ref{algo:dfs-decomp}.

\begin{algorithm}[t!]
\caption{SplitInterval}%
\label{algo:fnsplit}
\DontPrintSemicolon
\KwIn{
    current level $l_{\mathrm{current}}$, maximum level $l_{\mathrm{max}}$, splitting factor $d$, array $\mathtt{solvelevels}$
}
\KwOut{
    number of new intervals $d_{\mathrm{current}}$
}
\eIf{$\mathtt{solvelevels}$ is not empty}{
    $l_{\mathrm{avg}} \gets \lfloor \mathrm{Average}(\mathtt{solvelevels})\rfloor$\;
    $d_{\mathrm{current}} \gets d^{l_{\mathrm{avg}}-l_{\mathrm{current}}}$\;
    $l_{\mathrm{new}} \gets l_{\mathrm{avg}}$\;
}{
    \eIf{$l_{\mathrm{max}} > l_{\mathrm{current}}$}{
        $d_{\mathrm{current}} \gets d^{l_{\mathrm{max}}-l_{\mathrm{current}}}$\;
        $l_{\mathrm{new}} \gets l_{\mathrm{max}}$\;
    }{
        $d_{\mathrm{current}} \gets d$\;
        $l_{\mathrm{new}} \gets l_{\mathrm{current}}+1$\;
    }
}
\Return{$d_{\mathrm{current}}$}
\end{algorithm}

\subsection{Additional Heuristics and Final Algorithm}
To enhance the efficiency of the basic algorithm from Subsection~\ref{subsec:basicalg}, we introduce two key heuristics that preserve the $T(C)$ tree structure while optimizing its construction process and enabling parallel solving of subtasks. These heuristics reduce the time needed to find satisfying assignments and minimize computational resource usage through intelligent branching management and subtask prioritization.

\begin{algorithm}[t!]
\caption{Parallel CircuitSAT decomposition (master process)}%
\label{algo:par-decomp}
\DontPrintSemicolon
\KwIn{
    CNF formula $C$, number of initial intervals $q$, splitting factor $d$, timeout $t$
}
\KwOut{
    SAT (satisfying assignment) or UNSAT
}
Send $C$ to all workers\;
Initialize $\mathcal{R}_0^q = \{I_1, \dots, I_q\}$ and empty priority queue $Q$\;
\ForEach{$j \in \{1, \dots, q\}$}{
    Push $(I_j, 1)$ onto $Q$\;
}
Initialize $l_{\mathrm{max}} \gets 1$; $\mathtt{solvelevels} \gets \emptyset$\;
\While{($Q$ is not empty) \textbf{or} ($\mathtt{runningworkers}\neq\emptyset$)}{
    \tcp{Wait for message from worker}
    $\mathtt{message} \gets \mathtt{worker\_message}$\;
    \eIf{Got READY message from worker}{
        Pop $(I_{\mathrm{current}}, l_{\mathrm{current}})$ from $Q$\;
        $\mathtt{SendTaskToWorker}((I_{\mathrm{current}}, l_{\mathrm{current}}, t))$\;
        Add worker to $\mathtt{runningworkers}$\;
    }{
        \tcp{Got DONE message from worker}
        Get $\mathtt{result}$, $l_{\mathrm{current}}$, $I_{\mathrm{current}}$ from $\mathtt{message}$\;
        $l_{\mathrm{max}} \gets \max(l_{\mathrm{max}}, l_{\mathrm{current}})$\;
        \eIf{$\mathtt{result} = \mathrm{SAT}$}{
            Send EXIT signal to all workers.\;
            \Return{SAT (satisfying assignment)}
        }{
            \eIf{$\mathtt{result} = \mathrm{UNSAT}$}{
               Append $l_{\mathrm{current}}$ to $\mathtt{solvelevels}$\;
                \textbf{Continue} \tcp{Discard the current interval}
            }{
                \tcp{$\mathtt{result}$ is INDET}
                $d_{\mathrm{current}} = \mathtt{SplitInterval}$($l_{\mathrm{current}}$, $l_{\mathrm{max}}$, $d$, $\mathtt{solvelevels}$)\;
                Split $I_{\mathrm{current}}$ into $d_{\mathrm{current}}$ subintervals $\{I_1', \dots, I_{d_{\mathrm{current}}}'\}$\;
                \ForEach{$i \in \{1, \dots, d_{\mathrm{current}}\}$}{
                    Push $(I_i', l_{\mathrm{new}})$ onto $Q$\;
                }
            }
        }
        Remove worker from $\mathtt{runningworkers}$\;
    }
}
Send EXIT signal to all workers\;
\Return{UNSAT}\;
\end{algorithm}

\begin{heuristic}[Priority processing by tree level]
\label{heu1}
\normalfont
Each node of the tree $T(C)$ is assigned a level $l$ that reflects the depth of decomposition: the root has level $l=0$, the $q$ initial subtasks have level $l=1$, and each subsequent split increases the level by one. The subtasks are placed in a priority queue ordered by decreasing level $l$, which focuses processing on intervals of smaller size. These higher-level intervals (with larger $l$ values) contain fewer possible solutions to inequality \eqref{eq3}, thereby increasing the probability that the SAT solver $A^t$ can solve the subtask.
\end{heuristic}

\begin{algorithm}[t!]
\caption{Parallel CircuitSAT decomposition (worker process)}%
\label{algo:worker}
\DontPrintSemicolon
\KwIn{
    CNF formula $C$, SAT solver $A$
}
\While{no EXIT signal}{
    Send READY message to master\;
    \tcp{Wait for message from master}
    $(I_{\mathrm{current}}, l_{\mathrm{current}}, t) \gets \mathtt{master\_message}$\;
    $C_{I_{\mathrm{current}}} \gets \mathtt{EncodeIntervalToCNF}(C, I_{\mathrm{current}})$\;
    $\mathtt{result} \gets \mathtt{SolveCNF}(C_{I_{\mathrm{current}}}, A^t)$\;
    \tcp{$\mathtt{result}$ can be SAT, UNSAT, or INDET}
    Send DONE($\mathtt{result}, I_{\mathrm{current}}, l_{\mathrm{current}}$) to master\;
}
\end{algorithm}

\begin{heuristic}[Adaptive interval splitting]
\label{heu2}
    \normalfont
The number of new subtasks generated by splitting an interval depends on parameter $d$, current level of $T(C)$ $l_{\mathrm{current}}$, average solving level $l_{\mathrm{avg}}$, and maximum level $l_{\mathrm{max}}$ reached during execution. When the SAT solver proves SAT or UNSAT for a subtask, its level is added to array $\mathtt{solvelevels}$. If $|\mathtt{solvelevels}|>0$, we can compute $l_{\mathrm{avg}} = \lfloor \mathrm{average}(\mathtt{solvelevels}) \rfloor$.

For an interval $I$ at level $l_{\mathrm{current}}$, assume that $C \land C_I$ returns INDET. The number of subintervals $d_{\mathrm{current}}$ in the splitting of interval $I$ is determined by the following rules:
\begin{enumerate}
    \item If $|\mathtt{solvelevels}|>0$ and $l_{\mathrm{current}}<l_{\mathrm{avg}}$, then $d_{\mathrm{current}}=d^{l_{\mathrm{avg}}-l_{\mathrm{current}}}$. This enables reaching levels with higher solution probability while minimally increasing queue size.

    \item Else, if $l_{\mathrm{current}}<l_{\mathrm{max}}$, then $d_{\mathrm{current}}=d^{l_{\mathrm{max}}-l_{\mathrm{current}}}$, immediately adapting to the current maximum tree depth.

    \item Else, if $l_{\mathrm{current}}=l_{\mathrm{max}}$, then $d_{\mathrm{current}}=d$ and $l_{\mathrm{max}}$ is incremented by~$1$.
\end{enumerate}
\end{heuristic}

The pseudocode implementing Heuristic~\ref{heu2} appears in Algorithm~\ref{algo:fnsplit}.

These heuristics significantly reduce computational resources while preserving the basic algorithm's completeness and correctness.
They efficiently explore the search space by focusing on promising levels of the decomposition tree.

Based on these heuristics, we developed a parallel CNF-solving algorithm using MPI. The master process (Algorithm~\ref{algo:par-decomp}) manages the task queue, while worker processes (Algorithm~\ref{algo:worker}) solve individual tasks.

\section{Computational Experiments}
This section presents experimental results demonstrating the effectiveness of our adaptive CircuitSAT partitioning algorithm.

\subsection{Benchmarks}
We evaluated our approach on two benchmark classes. The first class comprises Logical Equivalence Checking (LEC) instances for algorithms that sort $k$ natural numbers represented by $l$-bit vectors. Specifically, we examined the following sorting algorithms: bubble sort, selection sort~\cite{cormen90}, and pancake sort~\cite{gates1979}. The corresponding tests are denoted as $\mathtt{BvS}_{k,l}$ (Bubble versus Selection), $\mathtt{BvP}_{k,l}$ (Bubble versus Pancake), and $\mathtt{PvS}_{k,l}$ (Pancake versus Selection) LEC problems.

The second benchmark class consists of CNF formulas encoding preimage attacks on the MD4 cryptographic hash function. We focus on attacks against a step-reduced variant of MD4's compression function. MD4~\cite{DBLP:conf/crypto/Rivest90}, one of the earliest practical cryptographic hash algorithms, uses the Merkle--Damg{\aa}rd construction~\cite{DBLP:conf/crypto/Merkle89,DBLP:conf/crypto/Damgard89a}. While vulnerable to collision attacks~\cite{DBLP:conf/eurocrypt/WangLFCY05} and now considered obsolete, no practical preimage attack exists even for its compression function. The best known attack~\cite{DBLP:conf/fse/Leurent08} requires $2^{96}$ function calls, making it impractical.

Realistic attacks target the compression function with reduced steps. The original 48-step algorithm's variants are denoted MD4-$k$ ($k\leq48$). The first successful attack on MD4-32 appeared in~\cite{DBLP:conf/fse/Dobbertin98}. Later, attacks for $k\leq39$ achievable on personal computers were published in~\cite{DBLP:conf/sat/DeKV07,DBLP:conf/mipro/GribanovaS18}. To our knowledge, the largest tractable variant is MD4-43, with attacks described in~\cite{DBLP:conf/ijcai/Zaikin2022,DBLP:journals/jair/Zaikin24,IEEE2025}. We apply our Section~\ref{sec4} algorithm to improve upon~\cite{IEEE2025}'s results.

\subsection{Computational Platform}
All experiments were conducted on the Academician V.~M.~Matrosov computing cluster at the Irkutsk Supercomputer Center of the Siberian Branch of the Russian Academy of Sciences~\cite{Matrosov}. The cluster's main module contains 60 nodes, each with two 18-core Intel Xeon E5-2695 v4 ``Broadwell'' processors. For our tests, we utilized between one (36 cores) and five (180 cores) nodes.

\subsection{Experimental Results}
The first series of experiments addressed LEC problems for sorting algorithms that the sequential solver Kissat~4.0.1~\cite{kissat-cadical} could not solve within 24 hours but which our algorithm successfully solved using one cluster node (36 cores). Table~\ref{tab:sort36} presents the results.

\begin{table}[t!]
\centering
\caption{Results of applying Algorithm~\ref{algo:par-decomp} to LEC instances for sorting algorithms (36-core wall-clock time)}
\label{tab:sort36}
\setlength{\tabcolsep}{0.4em}
\renewcommand{\arraystretch}{1.3}%
\footnotesize
\begin{tabular}{@{}ccrrccccc@{}}
 \toprule
 \thead{Instance} &
 \thead{q} &
 \thead{d} &
 \thead{t} &
 \thead{Number\\of\\INDETs} &
 \thead{Max.\\reached\\level} &
 \thead{CPU\\time (s)} &
 \thead{Wall-clock\\time (s)} &
 \thead{CPU/wall\\ratio}\\
    \hline
    $\mathtt{BvP}_{11,3}$ & 1000 & 3 & 500 & 25 & 2 & \ 332\,392 & \ \ 9698 & 34.2\\
    $\mathtt{BvP}_{17,2}$ & 1000 & 3 & 500 & 922 & 3 & 1\,333\,851 & 38\,523 & 34.6\\
    $\mathtt{PvS}_{9,4}$ & 1000 & 3 & 500 & 1548 & 3 & 2\,000\,801 & 57\,649 & 34.7\\
    $\mathtt{PvS}_{14,2}$ & 1000 & 3 & 500 & 0 & 1 & \ \,\ \ 73\,971 & \ \ 2170 & 34.0\\
    $\mathtt{PvS}_{11,3}$ & 1000 & 3 & 500 & 1251 & 3 & 1\,723\,511 & 49\,553 & 34.7\\
\bottomrule
\end{tabular}
\end{table}

Let us analyze the table contents. All experiments began with $q=1000$ initial partitions, constructed using the interval division scheme from~\cite{IEEE2025}, enabling the division of any interval of length greater than $1$ (in the sense of the definition given in Section~\ref{sec3}) into equal intervals of smaller length. The parameter $d$ determines how many subintervals to create when splitting an interval, while $t$ specifies the time limit (in seconds) for the SAT solver (Kissat 4.0.1) to solve one task at each decomposition level.

The instances in Table~\ref{tab:sort36} exceeded the 100\,000-second limit on a single Intel E5-2695 core. The table columns show
\begin{itemize}
    \item Number of INDETs: count of interrupted tasks requiring further splitting;
    \item Max. reached level: maximum depth in the $T(C)$ tree;
    \item CPU time: total runtime (36 or 180 cores);
    \item Wall-clock time: actual elapsed time (36 or 180 cores);
    \item CPU/wall ratio: parallel speedup factor for 36 cores (180 cores in Table~\ref{tab:sort180}).
\end{itemize}

For the second experiment series, we addressed harder LEC instances unsolvable within 100\,000 seconds on one node, utilizing five cluster nodes (180 cores). Table~\ref{tab:sort180} presents these results.

\begin{table}[t!]
\centering
\caption{Results of applying Algorithm~\ref{algo:par-decomp} to hard LEC instances (180-core wall-clock time)}
\label{tab:sort180}
\setlength{\tabcolsep}{0.4em}
\renewcommand{\arraystretch}{1.3}%
\begin{tabular}{@{}ccrrccccc@{}}
 \toprule
 \thead{Instance} &
 \thead{q} &
 \thead{d} &
 \thead{t} &
 \thead{Number\\of\\INDETs} &
 \thead{Max.\\reached\\level} &
 \thead{CPU\\time (s)} &
 \thead{Wall-clock\\time (s)} &
 \thead{CPU/wall\\ratio}\\
    \hline
    $\mathtt{BvP}_{12,3}$ & 1000 & 3 & 500 & 2858 & 4 & 4\,862\,491 & 27\,771 & 175.0\\
    $\mathtt{BvS}_{13,3}$ & 1000 & 3 & 500 & 2106 & 5 & 8\,478\,171 & 48\,015 & 176.5\\
\bottomrule
\end{tabular}
\end{table}

The third experiment series addressed inversion problems for MD4-$k$ cryptographic hash functions (described previously). We examined MD4-$k$ for $k\in\{40,41,42,43\}$, targeting the inversion of the hash $1^{128}$ (a 128-bit vector of ones). Initial parameters $q$ and $t$ were chosen similar to those in~\cite{IEEE2025} for MD4-40 and MD4-43 inversions.

Notably, the algorithm continued processing intervals even after finding satisfying assignments, stopping only when queue $Q$ emptied and all subtasks were completed. This ensured a fair comparison with results from~\cite{IEEE2025}. Table~\ref{tab:md4} presents these findings.

\begin{table}[t!]
\centering
\caption{Results of applying Algorithm~\ref{algo:par-decomp} to MD4 hash function inversion problems (180-core wall-clock time)}
\label{tab:md4}
\setlength{\tabcolsep}{0.4em}
\renewcommand{\arraystretch}{1.3}%
\begin{tabular}{@{}ccrrccccc@{}}
 \toprule
 \thead{Instance} &
 \thead{q} &
 \thead{d} &
 \thead{t} &
 \thead{Number\\of\\INDETs} &
 \thead{Max.\\reached\\level} &
 \thead{CPU\\time (s)} &
 \thead{Wall-clock\\time (s)}&
 \thead{CnC\\wall-clock\\time (s)}\\
    \midrule
    $\mathtt{MD4\text{-}40}$ & 60\,000 & 2 & 220 & 3089 & 4 & 6\,763\,298 & 43\,054 & ---\\
    $\mathtt{MD4\text{-}41}$ & 60\,000 & 2 & 220 & 7455 & 4 & 9\,128\,960 & 57\,660 & $>$300\,000\\
    $\mathtt{MD4\text{-}42}$ & 60\,000 & 2 & 220 & 1904 & 4 & 5\,858\,125 & 37\,498 & $>$300\,000\\
    $\mathtt{MD4\text{-}43}$ & 60\,000 & 2 & 220 & 4020 & 3 & 7\,027\,119 & 43\,288 & \ \ 264\,039\\
\bottomrule
\end{tabular}
\end{table}

The ``CnC wall-clock time'' column shows results from the \textup{Cube-and-Conquer} strategy~\cite{CC2012} with default settings. We generated cubes using \texttt{march\_cu}\footnote{\url{https://github.com/marijnheule/CnC}.} and solved them with Kissat~4.0.1. For MD4-40, \texttt{march\_cu} failed to construct cubes within 100\,000 seconds. For MD4-41 and MD4-42, solving exceeded 300\,000 seconds without finding solutions. MD4-43 required 264\,039 seconds to find a satisfying assignment.

Our algorithm demonstrated superior performance over both~\cite{IEEE2025} and default Cube-and-Conquer for MD4-40 through MD4-43 inversion problems.

\section{Conclusions and Future Work}

We presented a novel approach for solving hard SAT instances arising from CircuitSAT problems. The algorithm's key innovation involves employing a time-limited SAT solver within a SAT partitioning framework. When the solver cannot complete a task within its allotted time, the task is interrupted and decomposed into simpler subtasks using the interval partitioning scheme from~\cite{IEEE2025}. Our experimental results demonstrate the method's effectiveness on hard Logical Equivalence Checking (LEC) instances and preimage attacks against weakened variants of the MD4 compression function.

The algorithm shows considerable potential for further enhancement through several avenues. First, developing heuristics for automated selection and dynamic adjustment of parameters $q$, $d$, and $t$ could significantly improve performance. Second, sharing learned clauses between different branches of the decomposition tree $T(C)$ presents a promising optimization opportunity. We intend to investigate these directions in future work.

\subsection*{Acknowledgments.}
This research was financially supported by the Ministry of Education and Science of the Russian Federation (State Registration \textnumero\,\mbox{121041300065-9}).

We thank Stepan Kochemazov for his advice, which helped us improve the presentation of the manuscript.

%
%
%
\bibliographystyle{spmpsci}
\bibliography{refs}

\end{document}